%% file: main.tex
\documentclass[11pt, a4paper, logo, copyright, nonumbering]{map}
\input{resources/packages}

\definecolor{hidden-draw}{RGB}{20,68,106}
\definecolor{hidden-pink}{RGB}{255,245,247}
\usepackage[edges]{forest}

\usepackage{bm}

\usepackage{natbib}
\usepackage{CJKutf8}
\usepackage{xargs}
\usepackage{todonotes}
\usepackage{multirow}
\usepackage{cleveref}
\usepackage{subcaption}
\usepackage{url}
\usepackage{svg}
\usepackage{fontawesome}
\usepackage{xcolor}
\usepackage{float}

\usepackage[utf8]{inputenc}
\usepackage{booktabs}
\usepackage{graphicx}
\usepackage{array}
\usepackage{tabularx}
\usepackage{xcolor,colortbl}  
\definecolor{boxcolor}{HTML}{d92523} 
\definecolor{bulbcolor}{HTML}{e3b87f} 
\usepackage{url}
\usepackage{ragged2e}   
\usepackage{makecell} 

\usepackage{hyperref}       
\usepackage{url}            
\usepackage{booktabs}       
\usepackage{nicefrac}       
\usepackage{microtype}      
\usepackage{xcolor}         
\usepackage{graphicx}
\usepackage{longtable}
\usepackage{array}
\usepackage{pbox}
\usepackage{tabularx}
\usepackage{multirow}
\usepackage{wrapfig}
\usepackage{pifont}
\usepackage{lipsum}
\usepackage{subcaption}
\usepackage{enumitem}
\usepackage{tikz}
\usepackage{xstring}

\usepackage{color-edits}

\usepackage{booktabs}
\usepackage{multirow}
\usepackage[table]{xcolor}
\usepackage{tabularx}

\usepackage{parskip} 

\usepackage{threeparttable} 

\usepackage{algorithm}
\usepackage{algpseudocode}

\usepackage{xspace}
\usepackage{times}
\usepackage{latexsym}
\usepackage{booktabs}
\usepackage{fvextra}
\usepackage{graphicx}
\usepackage{subcaption} 
\usepackage[T1]{fontenc}
\usepackage{xcolor}
\usepackage[utf8]{inputenc}

\usepackage{microtype}

\usepackage{inconsolata}
\usepackage{amsmath}
\usepackage{booktabs}
\usepackage{graphicx}
\usepackage{pifont}
\usepackage{multirow}          
\usepackage{colortbl}          
\usepackage{algorithm}
\usepackage{algpseudocode}
\usepackage{xcolor}
\usepackage{amssymb}

\definecolor{rliableolive}{HTML}{BBCC33}
\definecolor{rliableblue}{HTML}{77AADD}
\definecolor{rliablered}{HTML}{f63c44}

\algrenewcommand\algorithmicrequire{\textbf{Input:}}
\algrenewcommand\algorithmicensure{\textbf{Output:}}
\algrenewcommand\alglinenumber[1]{\small #1:}

\newcommand{\eg}{\textit{e.g.,}\xspace}

\newcommand{\paratitle}[1]{\vspace{1.5ex}\noindent\textbf{#1}}

\usepackage{alltt}
\usepackage[most,skins,theorems]{tcolorbox}
\tcbuselibrary{skins,breakable,fitting,hooks}  
\tcbset{
  aibox/.style={
    width=\linewidth,
    top=7pt,
    bottom=8pt,
    colback=white,
    colframe=black,
    colbacktitle=black,
    enhanced,
    center,
    attach boxed title to top left={yshift=-0.1in,xshift=0.15in},
    boxed title style={boxrule=0pt,colframe=white},
  }
}
\tcbset{
  aibox2/.style={
    width=\linewidth,
    top=7pt,
    bottom=2pt,
    colback=green!18!white,
    colframe=black,
    colbacktitle=black,
    enhanced,
    center,
    attach boxed title to top left={yshift=-0.1in,xshift=0.15in},
    boxed title style={boxrule=0pt,colframe=white,},
  }
}
\newtcolorbox{AIbox}[2][]{aibox,title=#2,#1}
\newtcolorbox{AIbox2}[2][]{aibox2,title=#2,#1}

\usepackage{xcolor}

\hypersetup{
    colorlinks=true,            
    linkcolor=blue,             
    filecolor=magenta,          
    urlcolor=cyan,              
    citecolor=purple,             
    pdftitle={Overleaf Example},
    pdfpagemode=FullScreen,
}

\definecolor{iquestblue}{HTML}{173C7F}
\definecolor{iquestazure}{HTML}{528FCC}

\newcommandx{\info}[2][1=]{\todo[linecolor=red,backgroundcolor=red!25,bordercolor=red,#1]{#2}}

\title{
\vspace{-0.2in}
\centering \fontsize{15pt}{16pt}\selectfont

ClawGym II: Exploring Black-Box RL  on  Agent  Harness

\vspace{-0.2in}
}

\author{
Huatong Song\textsuperscript{1*},
Fei Bai\textsuperscript{1*},
Ming Yang\textsuperscript{2}, 
Renyuan Li\textsuperscript{2},
Jia Deng\textsuperscript{1},
Jujie He\textsuperscript{2},
Zhange Zhang\textsuperscript{2},\\
\normalfont
Daixuan Cheng\textsuperscript{1},
Yan Xing\textsuperscript{2},
Qi Yun\textsuperscript{2},
Xuxing Chen\textsuperscript{2},
Danyang Li\textsuperscript{2},
Feng Chang\textsuperscript{2},
Chuan Hao\textsuperscript{2},\\
\normalfont
Ran Tao\textsuperscript{2},
Jian Yang\textsuperscript{2},
Bryan Dai\textsuperscript{2},
Wayne Xin Zhao\textsuperscript{1$\dagger$},
Mingjie Tang\textsuperscript{2$\dagger$},
Ji-Rong Wen\textsuperscript{1}\\
\normalfont
\textsuperscript{1}Gaoling School of Artificial Intelligence, Renmin University of China, \\
\normalfont
\textsuperscript{2}IQuest Research \\
\normalfont
\normalsize{\textsuperscript{*}Equal contributors\quad
\textsuperscript{$\dagger$}Corresponding Authors \\
\normalfont
Email: \texttt{\{songhuatong123, feibai\}@ruc.edu.cn}, \\
\normalfont
\texttt{batmanfly@gmail.com}
}}

\newcommand{\ignore}[1]{}

\usepackage[most]{tcolorbox}
\definecolor{darkorange}{RGB}{255, 140, 0}
\definecolor{lightgreen}{RGB}{145, 204, 117}
\definecolor{lightyellow}{RGB}{250, 200, 88}
\definecolor{lightred}{RGB}{238, 102, 102}
\definecolor{lightblue}{RGB}{115, 192, 222}

\definecolor{gray_1}{HTML}{B7B7B7}
\definecolor{gray_2}{HTML}{F0F0F0} 
\definecolor{frame_blue}{HTML}{A9D18E}

\newtcolorbox[auto counter, number within=section]{PromptBox}[2][]{
    enhanced,
    breakable,
    colback=gray_2, 
    colframe=gray_1,
    coltitle=white,
    fontupper=\small,
    fonttitle=\bfseries,
    title={#2}, 
    label={#1},
    arc=2pt,
    boxrule=1pt,
    left=2mm, right=2mm, top=2mm, bottom=2mm,
}

\usepackage{etoolbox}
\makeatletter
\patchcmd{\@maketitle}
  {\@toptitlebar}
  {%
    \hbox to \textwidth{%
    
    \includegraphics[height=0.7cm]{figures/aibox-logo.png}%
      \hfill
      \includegraphics[height=0.7cm]{figures/logo2.pdf}%
    }%
    \vspace{0.3em}
    \@toptitlebar
  }
  {}{}
\makeatother

\input{sections/0_abstract}

\begin{document}
\maketitle

\let\oldthefootnote\thefootnote



\input{sections/1_introduction}

\input{sections/2_method}

\input{sections/3_exp}

\input{sections/5_conclusion}


\bibliography{ref}
\newpage
\appendix
\section{Evaluation Prompt}\label{app:prompt}

\begin{PromptBox}[p:judge_prompt]{Rubric-Based Judge}
\begin{lstlisting}
# System prompt:

You are a strict rubric-based evaluator. Use only the user prompt content. Do not call tools, browse, inspect files, or ask for more context. Your response may include concise analysis, but it must end with exactly one standalone JSON object with keys `scores` and `notes`.

# User prompt:

You are grading an OpenClaw agent result.

You must not call, request, or simulate any tools. Do not browse, list directories, open files, inspect the workspace, or ask for more context. Grade only from the task, final output files, optional transcript evidence, and rubrics included in this prompt.

Before giving the final judgment, provide concise analysis explaining how the final outputs satisfy or fail each rubric criterion.

Your final judgment must end with exactly one standalone JSON object and nothing else after it. The final JSON object must contain exactly two keys: `scores` and `notes`.

- `scores` must map every rubric id (`criterion_1`, `criterion_2`, ...) to one numeric score chosen from that rubric's allowed score anchors.
- `notes` must be a concise string summarizing the main reasons for the assigned scores.

Do not include any overall score in the final JSON. Do not compute or report the final aggregated score. Score aggregation will be handled separately by post-processing.

## User Task
{{USER_TASK}}

## Final Output Files
{{FINAL_OUTPUT_FILES}}

## Additional Changed Workspace Files
{{ADDITIONAL_CHANGED_WORKSPACE_FILES}}

## Transcript
{{TRANSCRIPT_OPTIONAL}}

## Rubric
{{RUBRIC}}

\end{lstlisting}
\end{PromptBox}



\end{document}

%% file: resources/packages.tex
\usepackage{CJKutf8}
\usepackage{xargs}  
\usepackage{todonotes}
\usepackage{amsmath}
\usepackage{dsfont}

\usepackage{longtable}

\usepackage{svg}
\usepackage{fontawesome}
\usepackage{array}
\usepackage{tabularx}
\usepackage{latexsym}
\usepackage{graphicx}
\usepackage[utf8]{inputenc} 
\usepackage[utf8]{inputenc} 
\usepackage{amssymb} 
\usepackage{url}            
\usepackage{amsfonts}       
\usepackage{nicefrac}       
\usepackage{microtype}      
\usepackage{xspace}

\usepackage{listings}
\usepackage{makecell}
\usepackage{inconsolata}
\usepackage{multicol}
\usepackage{amstext}

\definecolor{darkblue}{RGB}{84, 112, 198}

\definecolor{lightblue}{rgb}{0.85, 0.95, 1.0}    
\definecolor{lightgreen}{rgb}{0.90, 1.0, 0.90}    
\definecolor{lightorange}{rgb}{1.0, 0.95, 0.85}   
\definecolor{lightpurple}{rgb}{0.95, 0.90, 1.0}   
\definecolor{lightgray}{rgb}{0.97, 0.97, 0.97}    

\usepackage{tikz}
\usetikzlibrary{calc}
\definecolor{battery-empty}{rgb}{0.9, 0.9, 0.9}
\newcommand{\difficultybar}[1]{%
  \begin{tikzpicture}[baseline, scale=0.5, every node/.style={scale=0.8}]
    \foreach \i in {1,2,3,4,5} {
      \ifnum\i>#1
        \draw[fill=battery-empty] (\i*0.5-0.5, 0) rectangle (\i*0.5, 0.25);
      \else
        \pgfmathsetmacro{\colorlevel}{80 - 12*(\i)} 
        \edef\x{\noexpand\draw[fill=blue!\colorlevel!white, opacity=0.9] (\i*0.5-0.5, 0) rectangle (\i*0.5, 0.25);}
        \x
        \draw[blue!50!black] (\i*0.5-0.5, 0) rectangle (\i*0.5, 0.25);
      \fi
    }
    \fill[battery-empty!70] (2.5, 0.08) rectangle (2.6, 0.17);
    \draw[battery-empty!70!black] (2.5, 0.08) rectangle (2.6, 0.17);
  \end{tikzpicture}%
}
\renewcommand{\arraystretch}{0.96}

%% file: sections/0_abstract.tex
\begin{abstract}
Agent harnesses have substantially improved performance on long-horizon tasks by coordinating agent interactions with the environment.
However, reinforcement learning through complex harnesses remains largely unexplored, as scaling such training to long-horizon agent tasks introduces fundamental challenges. 
In this work, we present a \textbf{unified black-box RL framework} for stable and scalable optimization of general agents through complex harnesses.
 Concretely, we first build a sandbox-based execution infrastructure that isolates task environments and harnesses within temporary sandboxes for large-scale concurrent rollouts.
We then decouple policy optimization from opaque harness execution and place a serving proxy at the model boundary to capture model calls. To reconstruct multi-turn trajectories and improve training efficiency, we organize the captured calls into prefix trees  and further adapt both critic-based PPO and critic-free GRPO to optimize over the recovered tree structure. Meanwhile, we maintain training--inference consistency throughout the optimization process. Finally, we introduce \textbf{mix-harness training}, allowing a single model to be jointly optimized by heterogeneous harnesses. 
With Qwen3-30A3B, black-box RL improves Pass@1 on ClawGym-Bench by $9.98$ and $14.81$ points through \textbf{OpenClaw} and \textbf{Claude Code}, respectively, while remaining stable over 200--400 optimization steps.  
Moreover, the framework yields consistent gains on more challenging tasks such as JobBench and OfficeQA.
Overall, our framework enables effective, stable, and scalable optimization of general agents through black-box harnesses, supporting unified training across heterogeneous execution systems.

\end{abstract}

%% file: sections/1_introduction.tex
\section{Introduction}
\label{sec:introduction}

Agent harnesses~\citep{tang2026agent} have emerged as the foundational operating layer of autonomous agent systems, coordinating the interaction between large language models~\citep{zhao2023survey}  and their environments throughout the task-solving process.
They typically integrate system prompts, tool interfaces, context management, workflow orchestration, and retry and recovery mechanisms into a unified runtime. Production-grade systems such as Claude Code~\citep{anthropic2026claudecode}, Codex~\citep{openai2026codex}, and OpenClaw~\citep{openclaw2026} demonstrate that carefully designed harnesses can substantially improve agent performance on complex long-horizon tasks~\citep{rucnlpir2026awesomelonghorizonagents}.

As these harnesses become increasingly capable, they are accelerating the development of general autonomous agents that operate across diverse real-world workspaces and solve tasks spanning software engineering~\citep{merrill2026terminalbench, sweb-verified}, office applications~\citep{patwardhan2025gdpvalevaluatingaimodel, officeqa}, AI assistant workflows~\citep{ye2026claw, qwenclawbench1.1}, and MCP-enabled tool use~\citep{li2025tool, bandi2026mcp}. Recent progress has been driven by inference-side advances in harness design, optimization, and self-evolution~\citep{wang2026harness}. Yet the benefits of increasingly capable harnesses depend not only on their design but also on how effectively the underlying model can leverage them. Even a strong model may perform poorly within a capable harness if it has not been suitably trained, motivating efforts, typically via reinforcement learning~(RL) algorithms, to enable models to better exploit these agent harnesses. 

A defining characteristic of modern harnesses is their inherent complexity: the design and implementation are sufficiently intricate. As a result, RL training algorithms targeting a harness can differ substantially from those used in fully textual~\cite{chen2025empirical} or otherwise simple environments~\cite{song2025r1, song2026swe}. 
In simpler single-turn or lightweight agentic settings, the interaction process is explicit and directly observable, allowing RL to optimize well-defined model trajectories without accounting for complex hidden execution logic.
With modern harnesses, by contrast, the mainstream approach~\cite{xu2026polar, dressage_github} is to treat the internal control flow and execution logic as opaque and to design a {black-box RL} algorithm.   Indeed, black-box RL has become a key technical cornerstone for developing and leveraging capable harnesses for general agents~\cite{yu2026openforgerl}. Yet stable and scalable policy optimization for general agent tasks through such harnesses remains relatively underexplored in the existing literature. This work therefore focuses on the following research question: 
\begin{quote} \emph{How can black-box RL be performed through a harness to stably optimize general agents?} \end{quote}


To answer this question, we center our investigation on three fundamental challenges: 

$\bullet$ \emph{Scalable and stable infrastructure for black-box RL}. General agent tasks require dedicated, stateful environments during RL, and large-scale concurrent rollouts place substantial pressure on the execution infrastructure. For long-horizon tasks, runtime delays and failures can accumulate across interaction steps and invalidate entire trajectories, making timely and reliable data generation difficult at scale.

$\bullet$ \emph{Efficient and effective policy optimization from fragmented black-box traces}. Black-box harnesses expose fragmented, forked, and often redundant model calls rather than complete interaction trajectories, making these traces difficult to use directly for reinforcement learning. Moreover, opaque internal operations can further amplify training–inference inconsistency, leading to unstable training.

$\bullet$ \emph{Extensibility across heterogeneous harnesses}. Harnesses differ substantially in their interaction protocols, tool interfaces, context management, and execution workflows. A practical framework should support diverse harnesses with minimal system-specific adaptation and enable unified training across them, rather than optimizing policies for a single execution system.

To address these challenges, we develop a \textbf{unified black-box RL framework} to stably optimize general agents through any complex harness.
First, to sustain stable, large-scale concurrent environment execution during rollout, we build an execution and data infrastructure that preserves faithful model behavior.
Each task-specific environment and its execution harness are encapsulated within a temporary sandbox that is provisioned on demand and destroyed upon completion, providing an isolated workspace for every rollout while enabling concurrency at scale.
Second, we decouple policy optimization from harness execution by treating the harness as an opaque rollout engine, while a serving proxy intercepts every model call. This design allows the harness to retain its native execution logic and operate independently. To recover reliable multi-turn training trajectories and improve training efficiency, we organize the captured calls from each rollout into a \textbf{prefix tree}. We then adapt both critic-based PPO~\citep{schulman2017proximal} and critic-free GRPO~\citep{deepseekmath} to optimize over the recovered tree structure.
Training--inference consistency is further preserved through the \textbf{black-box token-in-token-out} mechanism~\citep{gallouedec2026tito} and token-level importance-sampling rollout correction~\citep{yao2025offpolicy}, which mitigate biases arising from differences between the underlying inference and training engines. Additional safeguards stabilize both rollout execution and policy optimization.
Finally, encapsulating each harness within the sandbox isolates harness-specific execution from the training pipeline, allowing different harnesses to be readily selected or replaced for different task environments.
Building on this modular design, we introduce \textbf{mix-harness training}, where each task--harness pair forms a basic training instance and rollouts from multiple heterogeneous harnesses are jointly used to optimize a single model.

We validate the black-box RL framework on two representative and structurally different harnesses: \textbf{OpenClaw}, a general-purpose harness for personal-assistant tasks, and \textbf{Claude Code}, a mature harness designed for long-horizon code and terminal interaction. General agent tasks from ClawGym~\citep{bai2026clawgym} are used for training under both harnesses. Starting from the Qwen3-30A3B~\citep{yang2025qwen3technicalreport} backbone, black-box RL training produces ClawII-OC-30A3B (trained through OpenClaw) and ClawII-CC-30A3B (trained through Claude Code), improving Pass@1 on PinchBench by $11.71$ and $17.28$ points and on ClawGym-Bench by $9.98$ and $14.81$ points, respectively. Additionally, training remains stable across {200--400 optimization steps}, demonstrating the effectiveness and robustness of our black-box RL framework for optimizing general agent tasks on complex harnesses. 
Furthermore, under mix-harness training, the trained model matches or even surpasses the performance of its counterparts trained with either harness alone, demonstrating that black-box RL can seamlessly integrate heterogeneous harnesses within our unified training pipeline. 
Beyond these settings, we extend our framework to  more challenging task settings from JobBench~\citep{li2026jobbenchaligningagentwork} and OfficeQA~\citep{opsahlong2026officeqaproenterprisebenchmark}, where black-box RL continues to deliver consistent improvements.

Our main contributions are summarized as follows:

\begin{itemize} 
\item We establish a unified and extensible black-box RL framework for general agents that accommodates training through any individual opaque harness and further enables mix-harness training via joint optimization across heterogeneous harnesses.

\item We develop the infrastructure and optimization techniques for stable and scalable black-box RL, supporting reliable concurrent rollouts, faithful trajectory recovery, tree-structure optimization with PPO and GRPO, and training–inference consistency.

\item We validate the framework on OpenClaw and Claude Code, demonstrating consistent performance gains and stable optimization, together with its extensibility across heterogeneous harnesses through mix-harness training and to more challenging tasks such as JobBench and OfficeQA.


\end{itemize}

%% file: sections/2_method.tex
\section{Preliminaries}
\label{sec:preliminaries}

\subsection{Problem Formulation for General Agent Tasks}
\label{subsec:task_formulation}

A general agent task consists of a user instruction together with an isolated, stateful environment in which the task is executed. The environment is initialized from a workspace and provides the execution context, including files, directories, documents, software repositories, running services, and other task-specific resources. To accomplish the task, the agent interacts with this environment over multiple turns through external tools, such as web search, file editing, code execution~\citep{bai2026towards}, shell commands, or API calls, progressively transforming its state toward the desired goal. The outcome is reflected in the final environment state, typically through generated artifacts (e.g., documents, presentations, or reports) or successfully completed operations (e.g., fixing a software repository or configuring a system). We denote a task as $q=(u,\mathcal{W}_0)$, where $u$ is the user instruction and $\mathcal{W}_0$ is the initial workspace, and use $\mathcal{E}_q$ to denote the associated execution environment initialized from $\mathcal{W}_0$, with tasks sampled from $q\sim\mathcal{D}$.

During execution, the agent alternates between observations and actions and produces a trajectory
\[
\tau=(o_1,a_1,\ldots,o_T,a_T),
\]
where $o_t$ denotes the observation at step $t$ and $a_t$ denotes the corresponding agent action. The interaction terminates when the task is completed or execution stops, resulting in a final workspace state $\mathcal{W}_T$. This final state is then evaluated to obtain a rollout-level reward.

In practice, task evaluation typically follows two paradigms. {Rule-based rewards} verify the correctness of the final workspace through deterministic procedures, such as unit tests, program execution, or exact-match validation. {Rubric-based rewards} rely on a stronger evaluator (e.g., an LLM-as-a-Judge) to assess the overall quality of the generated artifact or final workspace according to task-specific evaluation criteria.



\subsection{Harness-Driven Agent Execution}
\label{subsec:harness_driven_execution}

Modern general-purpose agents are increasingly deployed through mature harnesses, such as OpenClaw, Claude Code, and Codex. Compared with traditional handcrafted agent loops based on the standard ReAct~\citep{yao2022react} paradigm, these harnesses~\citep{qwenpaw2026, nanobot2026} provide a substantially richer execution abstraction between the language model and the external workspace. They integrate capabilities such as tool orchestration, context management, built-in skills, subagent delegation, and failure recovery into a unified runtime system, enabling agents to solve substantially more complex long-horizon tasks while abstracting away many low-level execution details from the model.

Under this execution paradigm, the model no longer interacts directly with the workspace. Instead, given a model action $a_t$, the harness executes the corresponding tools, updates the workspace and execution context, and returns the processed observation for the next decision. This interaction can be abstracted as

\begin{equation}
(\mathcal{W}_{t+1},\, o_{t+1},\, c_{t+1})
=
\mathcal{H}(\mathcal{W}_t,\, c_t,\, a_t),
\end{equation}

where $\mathcal{H}$ denotes the harness, $\mathcal{W}_t$ is the workspace state, $c_t$ is the maintained execution context, and $o_{t+1}$ is the observation returned to the model. Consequently, the interaction trajectory is determined jointly by the model and the harness rather than by the model alone, making the harness the fundamental execution abstraction for modern long-horizon agent systems.

\subsection{Agentic Reinforcement Learning Algorithms}
\label{subsec:agentic_rl_algos}

Two mainstream reinforcement learning paradigms are investigated, distinguished primarily by their reliance on an explicit value model: Proximal Policy Optimization (PPO) as a critic-based method, and Group Relative Policy Optimization (GRPO) as a critic-free method.
Throughout, we let the state $s_t$ denote the interaction history $(o_1, a_1, \dots, o_t)$ that precedes action $a_t$, so that the policy $\pi_\theta(a_t \mid s_t)$ and the value function $V_\phi(s_t)$ are both defined over this history.

\subsubsection{Proximal Policy Optimization (PPO)}
The PPO algorithm optimizes the policy model $\pi_\theta(a|s)$ in conjunction with an auxiliary value model $V_\phi(s)$ that estimates the state-value function. 
To exclude environment-generated and invalid tokens from policy optimization, a binary mask $M_t\in\{0,1\}$ is applied. The resulting masked clipping objective is
\begin{equation}
    \mathcal{L}_{\text{PPO}}(\theta)
    =
    \hat{\mathbb{E}}_t
    \left[
    M_t \cdot
    \min\left(
    \rho_t(\theta)\hat{A}_t,\,
    \text{clip}\bigl(\rho_t(\theta),1-\epsilon,1+\epsilon\bigr)\hat{A}_t
    \right)
    \right],
\end{equation}
where $\rho_t(\theta)=\frac{\pi_\theta(a_t\mid s_t)}{\pi_{\theta_{\text{old}}}(a_t\mid s_t)}$ denotes the token-level policy probability ratio, and $M_t=1$ for valid policy-generated tokens and $0$ otherwise.

The same mask is incorporated into Generalized Advantage Estimation (GAE) to prevent invalid positions from contributing to temporal credit assignment:
\begin{equation}
    \hat{A}_t
    =
    \sum_{l=0}^{\infty}
    (\gamma\lambda)^l
    \delta_{t+l}\cdot M_{t+l},
\end{equation}
where
$\delta_t=r_t+\gamma V_\phi(s_{t+1})M_{t+1}-V_\phi(s_t)$
denotes the temporal-difference residual. The critic network is optimized by minimizing the masked mean-squared loss:
\begin{equation}
    \mathcal{L}_{\text{Value}}(\phi)
    =
    \hat{\mathbb{E}}_t
    \left[
        M_t\cdot
        \left(V_\phi(s_t)-G_t\right)^2
    \right],
\end{equation}
where $G_t$ is the masked discounted return.

\subsubsection{Group Relative Policy Optimization (GRPO)}
To avoid the memory overhead and training complexity of a separate value model $V_\phi(s)$, GRPO estimates advantages relative to a group of sampled rollouts. For a given task $q$, the policy model $\pi_\theta$ generates a group of $G$ independent trajectories $\{\tau_1,\tau_2,\dots,\tau_G\}$ with corresponding rewards $\{R_1,R_2,\dots,R_G\}$. The advantage $\hat{A}_i$ of the $i$-th trajectory is calculated directly from the group rewards:
\begin{equation}
    \hat{A}_i
    =
    \frac{R_i-\mu_q}{\sigma_q+\varepsilon},
    \qquad
    \mu_q
    =
    \frac{1}{G}\sum_{j=1}^{G}R_j,
    \qquad
    \sigma_q
    =
    \sqrt{
        \frac{1}{G}
        \sum_{j=1}^{G}
        \left(R_j-\mu_q\right)^2
    }.
\end{equation}


Reusing the token-level mask and policy ratio introduced above, with an additional trajectory index, the masked GRPO objective is

\begin{equation}
\begin{aligned}
    \mathcal{L}_{\text{GRPO}}(\theta)
    =
    \frac{1}{G}
    \sum_{i=1}^{G}
    \frac{1}{\sum_{t=1}^{|\tau_i|}M_{i,t}}
    \sum_{t=1}^{|\tau_i|}
    M_{i,t}
    \Big[
        &\min\big(
            \rho_{i,t}(\theta)\hat{A}_i,\,
            \operatorname{clip}\big(
                \rho_{i,t}(\theta),
                1-\epsilon,
                1+\epsilon
            \big)\hat{A}_i
        \big)
        \\
        &-
        \beta\mathbb{D}_{\mathrm{KL}}
        \left(
            \pi_\theta(\cdot\mid s_{i,t})
            \Vert
            \pi_{\mathrm{ref}}(\cdot\mid s_{i,t})
        \right)
    \Big].
\end{aligned}
\end{equation}
Here, $M_{i,t}$ and $\rho_{i,t}$ are the trajectory-indexed counterparts of $M_t$ and $\rho_t$, respectively. $\pi_{\text{ref}}$ is the reference policy used to constrain policy drift, and $\mathbb{D}_{\text{KL}}$ is the Kullback--Leibler divergence.

\section{Unified Black-Box RL through Complex Agent Harnesses}
\label{sec:black_box_rl}

Our black-box RL formulation directly optimizes a trainable model through an unmodified and opaque agent harness, while decoupling policy optimization from harness execution. This section presents the execution infrastructure for scalable black-box rollouts, the recovery and optimization of training trajectories from captured model calls, and the extension to training across heterogeneous harnesses.

\subsection{Infrastructure for Scalable Black-Box RL Execution}
\label{subsec:bb_formulation}




Black-box RL for general agent tasks requires large-scale concurrent execution of dedicated, stateful environments. Unlike stateless generation, each rollout may modify files, processes, tool states, and other task-specific resources over many interaction steps, making isolated execution essential. For each task $q$, we initialize a task-specific environment from its initial workspace $\mathcal{W}_0$ and launch the selected harness inside a temporary sandbox. The sandbox provides an isolated workspace and the runtime dependencies required by the harness, preventing interference across concurrent rollouts. Each sandbox is provisioned at rollout start and released after completion, allowing execution to scale with available sandbox capacity.
In addition to isolating task states, each sandbox provides the external capabilities required during rollout. Beyond the basic tools supplied natively by the selected harness, common capabilities (\eg web search), together with task-specific tools, are exposed through Model Context Protocol (MCP) servers launched inside the sandbox. 
Standardized MCP interfaces ensure reliable access to these services throughout execution and allow new capabilities to be integrated without modifying the native harness workflow.

We decouple policy optimization from harness execution. The training engine is responsible for policy optimization, while the inference engine serves the current policy during rollout. The harness itself drives the interaction inside the sandbox and retains control over tool use, context management, retries, and environment interaction. This separation avoids reimplementing harness-specific control flow in the training pipeline and enables different black-box harnesses to be integrated with minimal adaptation. 

Although the internal control flow of the harness is inaccessible, model behavior remains observable at the model-serving boundary. Every model-generated action is produced through a model request that crosses this boundary. We therefore place a serving proxy at the boundary as the model endpoint used by the harness. For each request, the proxy invokes the current rollout policy, returns the generated response in the protocol expected by the harness, and records the exact input tokens, generated tokens, rollout log-probabilities, and associated task metadata. This boundary-level capture provides the model-side information required for optimization without instrumenting the internal harness logic.

\begin{figure}[t]
    \centering
    \includegraphics[width=0.95\linewidth]{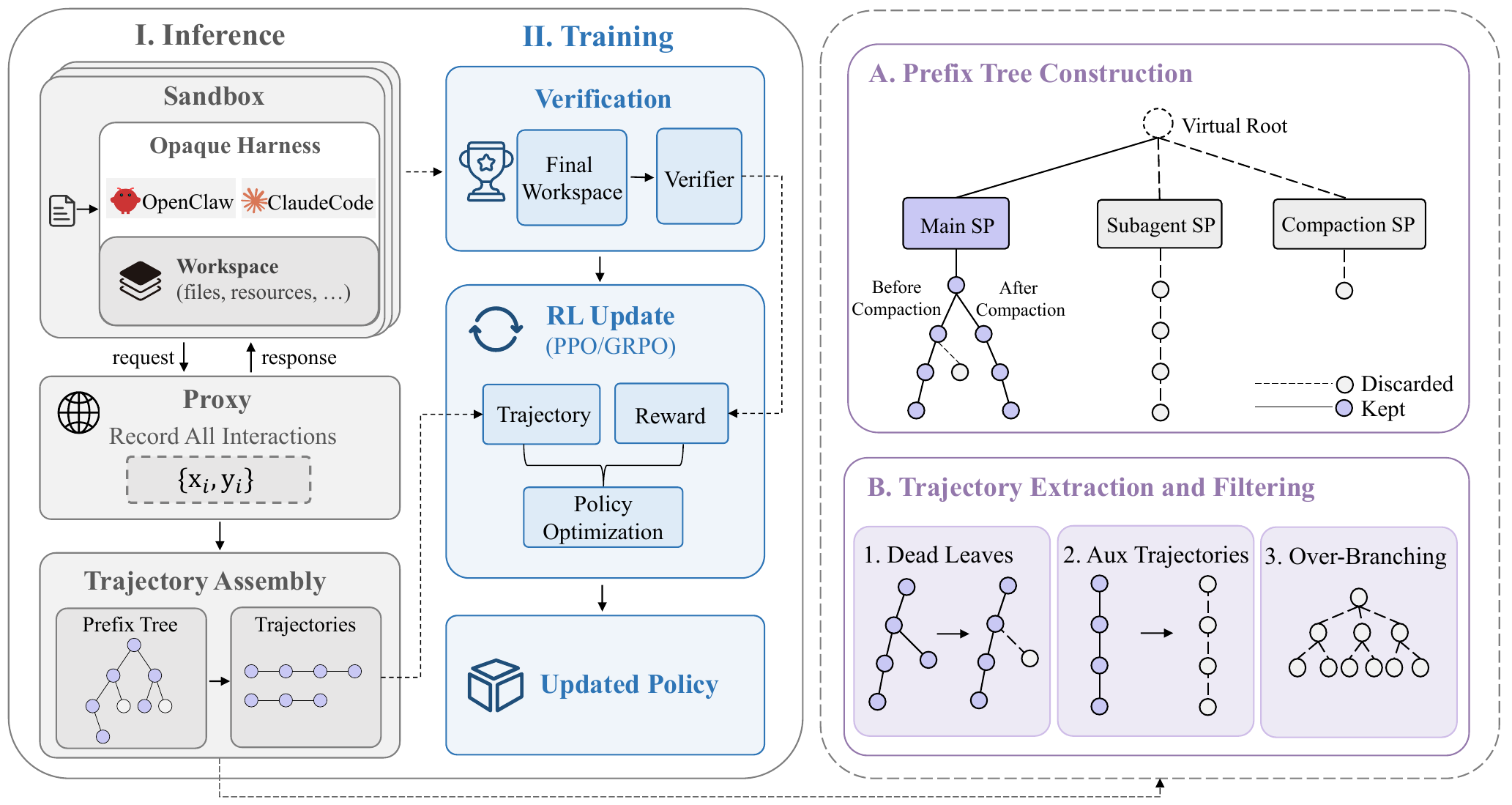}
    \caption{Overview of our  black-box RL framework for optimizing general agents through harnesses.}
    \label{fig:main}
\end{figure}

As illustrated in Figure~\ref{fig:main}, each rollout launches the task environment and harness inside a sandbox, after which all model requests pass through the serving proxy. Upon completion, the verifier evaluates the final workspace state and produces a rollout-level reward. The captured model-call records and reward are passed to the training pipeline, where the calls are reconstructed into trainable multi-turn trajectories, as described in Section~\ref{subsec:bb_bridge}. The training engine then updates the policy model and synchronizes the parameters to the inference engine for the next iteration.

\subsection{Bridging Black-Box Harness Execution and Policy Optimization}
\label{subsec:bb_bridge}



The serving proxy captures the model-side records required for optimization, but these records must first be transformed into structured multi-turn training trajectories. We organize the captured calls into prefix trees, optimize PPO and GRPO over the recovered tree structure, and preserve training--inference consistency throughout policy optimization.

\subsubsection{Prefix Tree Construction}\label{sec:tree_build}



The model calls captured during a black-box rollout are fragmented, forked, and potentially redundant.
Optimizing these calls independently would destroy the long-horizon interaction structure and repeatedly train on shared histories that recur across successive calls. We therefore organize the model calls captured from each rollout into a \emph{rollout-level prefix tree}~\citep{li2026kat}, which reconstructs their shared interaction structure and provides the training representation over which policy optimization is performed.

Concretely, consider a rollout that produces model calls
$\mathcal{C}=\{(x_i,y_i)\}_{i=1}^{m}$, where $x_i$ is the input context of the $i$-th call and $y_i$ is the corresponding model response. Within an uninterrupted interaction segment, later inputs extend histories established by earlier calls. 
Context compaction or subagent execution may start a new segment from a shortened or alternative context, after which subsequent calls continue to extend the newly established history.
We construct a tree $T$ rooted at the initial task prompt and attach each call to the existing node whose accumulated history forms the longest prefix of $x_i$. By comparing the input of a child call with the accumulated parent history and the recorded parent response, we also recover the intervening non-model content introduced by the harness, including tool outputs and other environment feedback.

A leaf is a call that no later call extends, and each root-to-leaf path defines a candidate multi-turn trajectory. Mature harnesses may introduce multiple leaves through context management or subagent scheduling, which start new interaction branches instead of continuing the current one.
By representing shared interaction histories with common tree nodes, the prefix tree stores each shared prefix only once while preserving distinct continuations as separate branches. The resulting leaves are processed as described in Section~\ref{subsubsec:bb_forked_extraction}, and the filtered tree is subsequently used for policy optimization as detailed in Section~\ref{sec:rl_algo_for_more_traj}.

\subsubsection{Trajectory Extraction and Filtering}
\label{subsubsec:bb_forked_extraction}


The prefix tree represents the model-call structure exposed during the task-solving process, and tracing a valid leaf back to the root recovers a candidate multi-turn training trajectory. However, not every leaf corresponds to the completion of the main task-solving process, as some arise from fault handling, context management, or subagent processes. We filter the resulting trajectories to retain those suitable for training.


\paratitle{Removing the Dead Leaves.}
A single turn may be generated multiple times when the inference server retries a request or the harness regenerates an invalid response, as OpenClaw does after a malformed tool call. Superseded attempts receive no further continuation but remain in the prefix tree as dead leaves. We partition the candidate trajectories into interaction segments delimited by context-compaction points and, within each segment, retain the leaf with the longest valid continuation. Shorter sibling trajectories terminating at retry-induced dead leaves are discarded.

\paratitle{Discarding Over-Branching Tasks.}
When the prefix tree of a rollout branches into an excessive number of leaves, the rollout has typically entered repeated or failed generation rather than legitimate interaction branching, and the resulting trajectories provide little useful training signal. Once the number of leaves exceeds a preset threshold, we discard the corresponding rollout, sacrificing a small amount of data to prevent corrupted signals from entering the gradient update.

\paratitle{Excluding Auxiliary Trajectories.}
Subagent and compaction trajectories are excluded because their roles in the harness differ from the role of the main task-solving trajectory and their contributions are not directly aligned with the rollout-level task reward. Broadcasting the same terminal reward to these auxiliary interactions would introduce ambiguous credit assignment and noisy optimization signals, potentially destabilizing training.
We therefore optimize only the main-agent trajectories and leave effective use of auxiliary interactions to future work.

\subsubsection{Policy Optimization over the Tree Structure}\label{sec:rl_algo_for_more_traj}





Once the prefix tree is constructed and invalid leaves are filtered, a single rollout may retain multiple valid root-to-leaf trajectories that share portions of the interaction history. Rather than collapsing them into a single trajectory, we retain this multi-trajectory structure and optimize the policy over the recovered tree. Since each rollout is evaluated once based on its final workspace state, all trajectories recovered from the same rollout inherit the same terminal reward. 
During optimization, all retained trajectories jointly contribute to the policy loss over the recovered tree. Trainable token nodes on branch-specific continuations are included normally, while nodes in shared prefixes are counted only once per rollout, preventing more highly branched rollouts from receiving disproportionate optimization weight.



Formally, we denote by $\{g_1,\dots,g_n\}$ the $n$ rollouts of a task $q$, where rollout $g_i$ contains trajectories $\{\tau_{i,1},\dots,\tau_{i,m_i}\}$ and is assigned a terminal reward $R_i$ shared across them. We next describe how GRPO and PPO are adapted to optimize over this multi-trajectory structure. The key adaptation is to preserve the rollout-level reward semantics while allowing all reconstructed trajectories from the same black-box rollout to contribute to policy optimization.

\paratitle{Advantage Estimate for GRPO.} 
GRPO naturally accommodates this structure: it estimates advantages from a group of rollouts for the same task and does not require a value model. We take the group to be the $n$ rollouts of $q$ and normalize each rollout's reward against the group statistics,

\begin{equation}
\hat{A}_i \;=\; \frac{R_i - \mu_q}{\sigma_q + \varepsilon},
\qquad
\mu_q = \frac{1}{n}\sum_{j=1}^{n} R_j,
\qquad
\sigma_q = \sqrt{\frac{1}{n}\sum_{j=1}^{n}\left(R_j - \mu_q\right)^2},
\end{equation}
where $\varepsilon$ guards against division by zero. 
The advantage $\hat{A}_i$ is computed once per rollout and assigned to all trainable token nodes covered by the retained trajectories in $g_i$, with tokens in shared prefixes contributing to the loss only once.

\paratitle{Advantage Estimate for PPO.} 
PPO employs an additional value model $V_\phi$ to estimate token-level advantages and jointly optimizes the policy and value models.
Fully modeling the dependencies among forked trajectories would require more complex credit assignment over the recovered tree structure. 
We therefore adopt a simplified variant of PPO in which trajectories within the same rollout are treated independently, with $\gamma=1$ and $\lambda=1$.



Specifically, each trajectory $\tau_{i,j}$ receives the reward $R_i$ assigned to its rollout at its own final token and performs GAE backup independently. 
No advantage signal is propagated across branch points connecting sibling trajectories. Under this setting, the standard GAE recursion degenerates to
\begin{equation}
\hat{A}_t = R_i - V_\phi(s_t),
\end{equation}
where the advantage is only related to the final reward and the value estimate at the current state. 
This simplification requires the value model to estimate long-horizon returns from intermediate states and may increase advantage variance because no temporal discounting is applied. 
A more principled treatment of forked trajectories is left for future work.

\subsubsection{Training–Inference Consistency}

During RL, the policy model must be trained on the exact token sequence generated during rollout, so that the sequence presented to the training engine matches the one actually sampled by the policy.
In the black-box setting, this requirement is complicated by harness-side transformations, such as tool-call normalization and assistant-message re-serialization, which may alter the representation of model outputs before they re-enter subsequent interactions.
Simply re-tokenizing the reconstructed root-to-leaf trajectories would encode the harness-transformed model responses, producing token sequences that may differ from those originally sampled during rollout.

To keep the trained sequence identical to the generated one, we adopt the \emph{black-box token-in-token-out} discipline~\citep{gallouedec2026tito} that maintains two decoupled views of every model call. 
The tokens generated by the inference engine are grafted directly onto the prefix tree and constitute the sole source of training data while the structured text that the harness receives is decoded from those same tokens purely for the harness to act on, and is never encoded back into the trained trajectory. Once a tool call is completed, the resulting environment response is encoded into a token sequence exactly as presented in the next model request and appended to the prefix tree.
The two views thus serve different consumers: the harness drives the interaction from the decoded text, while training reads the generated tokens verbatim. 
Regardless of how the harness reformats or standardizes what it displays and feeds into later turns,
those edits stay on its own view and cannot perturb the token record, so the sequence handed to the training engine is by construction identical to the one the policy model sampled.

Exact token sequences are insufficient, since the probability under which a token was sampled can differ from the one attributed to it at training time. 
The inference engine realizes the rollout policy $\pi_{\text{rollout}}$
and records the log probability of each sampled token, whereas the training engine recomputes token probabilities in a separate forward pass, denoted $\pi_{\text{old}}$, under different numerical precision, kernels, and parallelization. Even for the same token under the same prefix, the two probabilities can differ: 
\begin{equation}
\pi_{\text{rollout}}(a_t \mid s_t) \;\neq\; \pi_{\text{old}}(a_t \mid s_t).
\end{equation}

Ignoring this mismatch introduces off-policy bias into the gradient estimator~\citep{liu-li-2025-rl-collapse}.  To mitigate this remaining mismatch, we apply importance-sampling rollout correction~\citep{yao2025offpolicy}. In principle, a sequence-level importance ratio~\citep{liu-li-2025-rl-collapse} yields an unbiased estimator but can exhibit high variance over long trajectories. Since the mismatch between $\pi_{\text{rollout}}$ and $\pi_{\text{old}}$ is typically small, we adopt a token-level importance-sampling ratio, trading estimation bias for lower variance. Specifically, the loss of each training token is scaled by


\begin{equation}
w_t \;=\; \min\!\big( \exp\!\big( \log \pi_{\text{old}}(a_t \mid s_t) - \log \pi_{\text{rollout}}(a_t \mid s_t) \big),\; \bar{c} \big),
\end{equation}

where $\log \pi_{\text{rollout}}$ is recorded by the inference engine at rollout time, $\log \pi_{\text{old}}$
is recomputed by the training engine before policy updates, and the truncation threshold $\bar{c}$ bounds the variance from outlier ratios.

\subsection{Mix-Harness Training}
\label{subsec:bb_mix}

Harnesses can differ substantially in their interaction protocols, tool interfaces, context-management strategies, and control flows. Because our framework connects each harness to the same model-serving boundary and recovered trajectory representation, different harnesses can be readily integrated into a unified training pipeline without modifying the underlying policy optimization procedure. This design supports direct and flexible training through any individual harness under a common interface.

Training through a single fixed harness, however, may encourage the policy to specialize to its tool-use conventions, context-management strategy, and execution workflow. Such specialization may limit broader generalization when the same task is executed through another harness with different interaction patterns. We therefore introduce \emph{mix-harness training}, which jointly optimizes a shared policy using rollouts from multiple heterogeneous harnesses within the same training run.

For each task $q$ with environment $\mathcal{E}_q$, we construct separate training instances $(q,\mathcal{E}_q,H_k)$ by pairing the task with different compatible harnesses $H_k$. Since the environment is uniquely determined by the task, we refer to each instance as a task--harness pair. Instances associated with different harnesses are randomly mixed within each training batch and jointly processed through the shared training pipeline.

The main additional consideration is rollout grouping. For group-based advantage estimation, rollouts within a group must remain comparable. We therefore define each optimization group by a task--harness pair rather than by the task alone. Rollouts of the same task under different harnesses may appear in the same batch, but their advantages are normalized within separate task--harness groups. This prevents harness-dependent interaction patterns and reward distributions from distorting relative advantage estimation, while allowing all groups to jointly update the shared policy.




We instantiate both individual-harness and mix-harness training using two representative harnesses, OpenClaw and Claude Code.
OpenClaw is a general-purpose AI assistant harness supporting diverse workspace-grounded tasks, while Claude Code is a mature harness designed for long-horizon coding and terminal interaction. Detailed empirical results are provided in Section~\ref{sec:mix-harness-train}.

\subsection{Safeguards for Reliable Training}
\label{subsec:bb_stability}

Running RL through real harnesses and remotely provisioned sandboxes introduces failures and irregularities that arise from the execution infrastructure rather than the policy model itself. We employ several additional engineering safeguards to improve the reliability of rollout collection and training.

\paratitle{Robust Sandbox Execution and Fault Handling.}
Black-box RL relies on remotely provisioned sandboxes to execute real harnesses, making infrastructure-level failures such as execution timeouts, connection losses, and transient platform errors unavoidable yet unrelated to the policy model itself. If left unhandled, these failures can stall rollout collection or introduce invalid trajectories into training. To improve training robustness, we employ several fail-safe mechanisms. 
Long-horizon tasks are forcibly terminated once a wall-clock timeout is reached, while those that encounter explicit sandbox failures, such as HTTP errors or  initialization failures, are  discarded and excluded from both advantage estimation and policy optimization.

\paratitle{Buffered Pseudo-Streaming Parsing.}
Most black-box harnesses interact with agents through streaming generation and rely on incremental tool-call parsers to recover structured tool invocations online. In rare cases, however, incremental parsing may prematurely terminate a tool call because of parsing errors, even when the model generates a correct response. To address this issue, we adopt a \emph{pseudo-streaming} parsing scheme. During generation, output tokens are buffered while synthetic Server-Sent Events (SSE) are continuously emitted to keep the streaming connection alive. After the entire turn completes, the buffered output is parsed once using a non-streaming parser, eliminating incremental parsing errors while preserving the streaming interface.

\paratitle{Complete Trajectory Capture via Settling.}
Since black-box harnesses execute asynchronously and may perform retries or fault handling in the background, the completion or failure of an external request does not guarantee that all trajectory records have been written. Collecting them immediately may therefore produce incomplete trajectories. Before trajectory assembly, we wait until the number of records remains unchanged across several checks and no records are pending. A fixed timeout prevents indefinite waiting.

%% file: sections/3_exp.tex
\section{Experiments}
\label{exp}

We evaluate the effectiveness, stability, and extensibility of our black-box RL framework. We first report overall performance under OpenClaw and Claude Code, then examine training dynamics across harnesses and optimization algorithms and evaluate mix-harness training. 
We further extend the framework to more challenging tasks from JobBench and OfficeQA, study the effect of cold-start initialization, and compare black-box RL with a white-box AgentLoop setting.

\subsection{Overall Performance}

\subsubsection{Experimental Settings}

\paratitle{Evaluation Datasets and Metrics.}
We evaluate all models using Pass@1 on ClawGym-Bench~\citep{bai2026clawgym} and PinchBench~\citep{pinchbench2026} under a hybrid protocol that combines code-based verification with rubric-based judgment.
On ClawGym-Bench, tasks whose verifier consists solely of code checks are scored by that verifier alone, whereas tasks whose verifier combines code checks with rubrics are scored as a weighted sum of the two, with weights $0.7$ and $0.3$, respectively. For PinchBench, we take the task set released on April 10, 2026, discard the multimodal tasks, and retain $30$ tasks, each scored with the verifier provided by the benchmark. 
All rubric-based judgments use GPT-5.4~\citep{openai2026gpt54}, with the prompt in Appendix~\ref{app:prompt}.

\paratitle{Evaluated Models.}
We take as baselines the Qwen3 series~\citep{yang2025qwen3technicalreport} and ClawGym-Agents~\citep{bai2026clawgym}.
For black-box RL, we train Qwen3-8B and Qwen3-30A3B under two representative harnesses, OpenClaw~\citep{openclaw2026} and Claude Code~\citep{anthropic2026claudecode}, resulting in two model families, denoted as ClawII-OC and ClawII-CC, respectively. The training tasks are drawn from ClawGym-SynData~\citep{bai2026clawgym}. 
Due to data availability, ClawII-OC is initialized from ClawII-Cold, a model family obtained through a lightweight cold start on ClawGym-SynData, whereas ClawII-CC is trained directly from the corresponding base models.
All models are evaluated through black-box rollouts using the corresponding harness. The maximum context window is set to 64K tokens, with longer trajectories handled by the context management of the corresponding harness.

\subsubsection{Results}

\begin{table*}[ht]
\small
\centering

\caption{
Performance comparison of different models on ClawGym-Bench and PinchBench. The best and second-best Pass@1 results within each model group are highlighted in \textbf{bold} and \underline{underlined}.
}
\footnotesize
\setlength{\tabcolsep}{3.2pt}
\renewcommand{\arraystretch}{1.12}
\newcommand{\NA}{--}

\resizebox{0.95\textwidth}{!}{
\begin{tabular}{@{}l@{\hspace{2pt}}cccccccc@{}} 

\toprule
\multirow{2}{*}[-0.5ex]{\textbf{Model}} 
& \multirow{2}{*}[-0.5ex]{\makecell{\textbf{Pinch-}\\ \textbf{Bench}}}
& \multicolumn{7}{c}{\textbf{ClawGym-Bench}} \\
\cmidrule(lr){3-9}
& 
& \makecell{\textbf{Product.}\\\textbf{\& Collab.}} 
& \makecell{\textbf{Systems}\\\textbf{\& Auto.}} 
& \makecell{\textbf{Analysis}\\\textbf{\& Reason.}} 
& \makecell{\textbf{Content}\\\textbf{\& Domain}} 
& \makecell{\textbf{Planning}\\\textbf{\& Knowl.}} 
& \makecell{\textbf{Software}\\\textbf{Dev.}} 
& \textbf{Avg.} \\
\midrule
\rowcolor{gray!12}
\multicolumn{9}{c}{\textbf{OpenClaw as Harness}} \\
\addlinespace[2pt]
Qwen3-8B
& 54.50 & 37.46 & 29.06 & 30.40 & 41.12 & 44.47 & 30.69 & 35.02 \\
Qwen3-32B
& 49.40 & 40.68 & 36.21 & 37.84 & 47.16 & 49.29 & 33.11 & 40.32 \\
Qwen3-30A3B
& 55.60 & 42.47 & 42.09 & 45.04 & 51.95 & 45.98 & 46.24 & 45.11 \\
Qwen3-235A23B
& 60.60 & \underline{53.66} & {52.27} & 47.18 & 61.39 & \textbf{67.33} & 49.23 & 54.48 \\

ClawGym-8B
& 75.70 & 49.47 & 46.83 & 46.35 & 55.37 & 52.29 & 54.78 & 50.24 \\
ClawGym-30A3B
& \underline{86.00} & 52.98 & 50.97 & \textbf{64.64} & \underline{61.46} & 57.90 & \underline{56.13} & \underline{56.82} \\
ClawII-Cold-8B
&71.29&48.80&42.49&45.06&51.39&54.74&42.47&47.06 \\
ClawII-Cold-30A3B
&75.61&54.34&47.61&52.58&53.42&60.64&48.88&52.64 \\
\textbf{ClawII-OC-8B}
& 77.44 & 53.17 & \underline{55.99} & 51.28 & 59.60 & {61.68} & 50.83 & 54.98 \\
\textbf{ClawII-OC-30A3B}
& \textbf{87.32} & \textbf{62.75} & \textbf{59.81} & \underline{64.09} & \textbf{62.18} & \underline{63.12} & \textbf{66.71} & \textbf{62.62} \\
\midrule

\rowcolor{gray!12}
\multicolumn{9}{c}{\textbf{Claude Code as Harness}} \\
\addlinespace[2pt]
Qwen3-8B
& 27.40 & 16.19 & 13.94 & 20.26 & 25.37 & 23.62 & 14.65 & 18.54 \\
Qwen3-32B
& 53.27 & 33.34 & 25.63 & 27.68 & 33.03 & 27.85 & 28.30 & 29.37 \\
Qwen3-30A3B
& 54.14 & 38.75 & 33.06 & 32.99 & 45.43 & 41.62 & 32.26 & 37.06 \\
Qwen3-235A23B
& \underline{62.47} & \underline{41.74} & \underline{40.53} & \underline{43.70} & \underline{50.35} & \underline{49.03} & \underline{54.90} & \underline{45.59} \\
\textbf{ClawII-CC-8B}
& 61.21 & 41.27 & 38.19 & 41.65 & 49.33 & 44.61 & 39.37 & 42.05 \\
\textbf{ClawII-CC-30A3B}
& \textbf{71.42} & \textbf{47.84} & \textbf{45.59} & \textbf{48.88} & \textbf{57.05} & \textbf{56.93} & \textbf{62.93} & \textbf{51.87} \\

\bottomrule
\end{tabular}
}

\label{tab:main_results_by_category}
\end{table*}

Table~\ref{tab:main_results_by_category} shows the performance of our black-box RL framework. We make the following observations:

\noindent $\bullet$ \emph{Black-Box RL Delivers Substantial Gains.}
Black-box RL substantially improves the models used to initialize training. With the 30A3B backbone, ClawII-OC and ClawII-CC outperform their respective initial policies by 9.98 and 14.81 points on ClawGym-Bench under OpenClaw and Claude Code, respectively. Under OpenClaw, ClawII-OC-30A3B further exceeds the ClawGym-30A3B SFT baseline by 5.80 points. Consistent gains on PinchBench further show that these improvements transfer to an external benchmark.

\noindent $\bullet$ \emph{Consistent Gains across Heterogeneous Harnesses.}
Black-box RL remains effective across two structurally distinct harnesses, OpenClaw and Claude Code. When each model is trained and evaluated within the corresponding harness, it improves the 8B and 30A3B backbones by $7.92$ and $9.98$ points on OpenClaw, and by $23.51$ and $14.81$ points on Claude Code. The resulting 30A3B models further outperform Qwen3-235A23B by $8.14$ and $6.28$ points in the two settings, demonstrating the framework's applicability across heterogeneous harnesses.

\noindent $\bullet$ \emph{Robustness to Initialization Strategy.}
Black-box RL succeeds both with and without a cold-start stage. ClawII-OC is initialized from ClawII-Cold, whereas ClawII-CC is optimized directly from the corresponding base models. Both families improve consistently across model scales and task categories, indicating that a specialized warm-up stage is not required for effective training. Nevertheless, under the OpenClaw harness, introducing a lightweight cold start can provide a stronger initialization and lead to further performance gains. The analysis of the effect of cold-start initialization is provided in Section~\ref{sec:cold-start}.

\subsection{Training Dynamics Across Harnesses}




We study the training dynamics of black-box RL to examine whether the proposed framework supports stable policy optimization across heterogeneous harnesses and RL algorithms.
Using Qwen3-30A3B, we train through both OpenClaw and Claude Code with critic-based PPO and critic-free GRPO. As described in Section~\ref{sec:rl_algo_for_more_traj}, both algorithms are adapted to optimize over the recovered multi-trajectory tree structure. 


All experiments use tasks from ClawGym-SynData, with periodic evaluation on ClawGym-Bench. Due to the data availability constraint described earlier, the OpenClaw runs are initialized from a cold-started policy model, whereas the Claude Code runs begin directly from the corresponding base model. GRPO uses batches of $32$ tasks with $8$ rollouts per task, while PPO uses batches of $256$ tasks with $1$ rollout per task, resulting in the same rollout budget per update. Under this matched rollout budget, PPO covers a larger number of unique tasks per update, whereas GRPO allocates multiple rollouts to each task for group-based advantage estimation. PPO additionally requires a value model, 
which is initialized through a dedicated value-pretraining stage before RL optimization.

\begin{figure}[ht]
\centering
    \centering
    \includegraphics[width=0.95\linewidth]{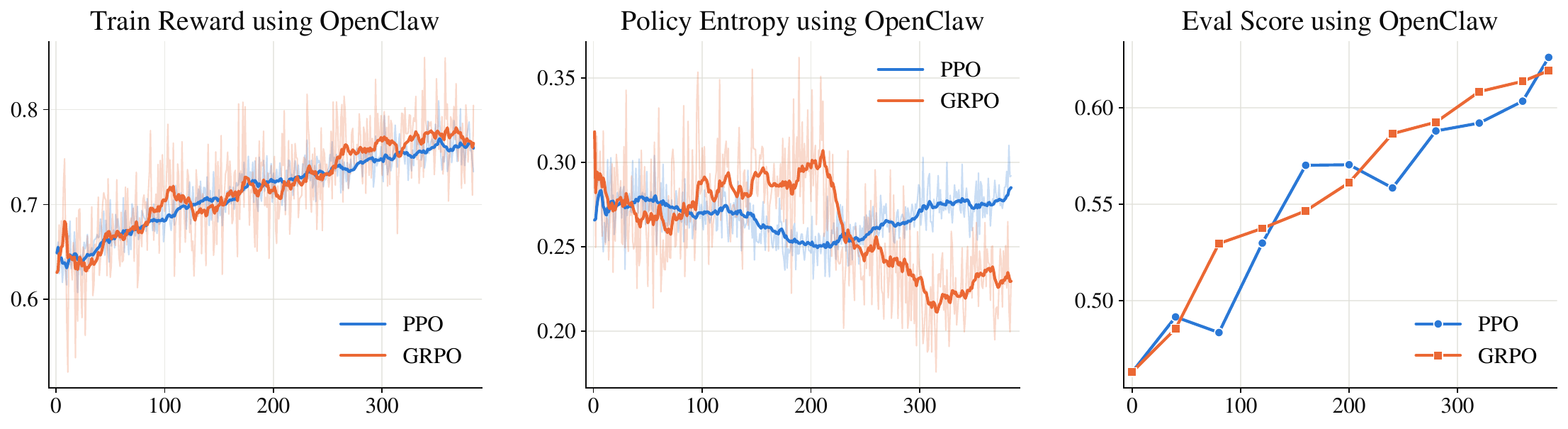}
\caption{Training dynamics of PPO and GRPO with OpenClaw as the rollout harness.}
    \label{fig:openclaw-training-dynamics}
\end{figure}

\begin{figure}[ht]
\centering
    \centering
    \includegraphics[width=0.95\linewidth]{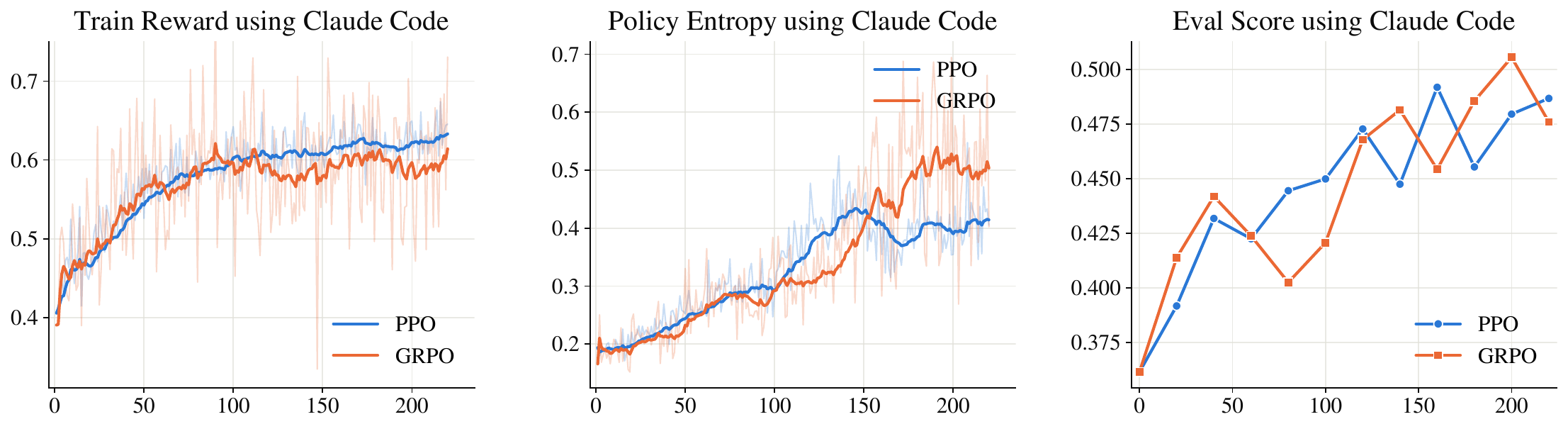}
\caption{Training dynamics of PPO and GRPO with Claude Code as the rollout harness.}
    \label{fig:claude-code-training-dynamics}
\end{figure}


As shown in Figures~\ref{fig:openclaw-training-dynamics} and~\ref{fig:claude-code-training-dynamics}, both PPO and GRPO maintain stable optimization over approximately 200--400 steps, showing clear upward trends in training reward and downstream evaluation performance while achieving broadly comparable final results.
PPO generally exhibits smoother entropy dynamics, whereas GRPO shows larger variation, including a late-stage entropy decline under OpenClaw.
Under Claude Code, both algorithms exhibit higher entropy than under OpenClaw before stabilizing. This difference may partly stem from the absence of cold-start initialization in the Claude Code runs, rather than from the harness itself.
We examine this effect in Section~\ref{sec:cold-start}. These results demonstrate that our black-box RL framework can reliably optimize general agent models through heterogeneous harnesses under both critic-based PPO and critic-free GRPO.


\subsection{Mix-Harness Training}\label{sec:mix-harness-train}


We further investigate whether a single model can be jointly optimized through multiple heterogeneous harnesses within a unified black-box RL pipeline. Starting directly from Qwen3-30A3B without cold-start initialization, we perform GRPO training with both OpenClaw and Claude Code. 
As described in Section~\ref{subsec:bb_mix}, for each task, we instantiate two task--harness pairs by executing the same task environment through OpenClaw and Claude Code, respectively. Task--harness pairs from both execution systems are randomly mixed within the same training batch, while GRPO grouping and relative-advantage normalization are performed independently for each pair. Thus, rollouts generated through different harnesses can coexist in the same batch but do not share the same group statistics or advantage baseline. The resulting gradients from all task--harness groups are aggregated within the same optimization step to jointly update a single policy model. Both the individual-harness and mix-harness settings use the same rollout batch configuration of 32 task--harness instances with 8 rollouts per instance, with periodic evaluation on ClawGym-Bench.

\begin{figure}[ht]
\centering
    \centering
    \includegraphics[width=0.95\linewidth]{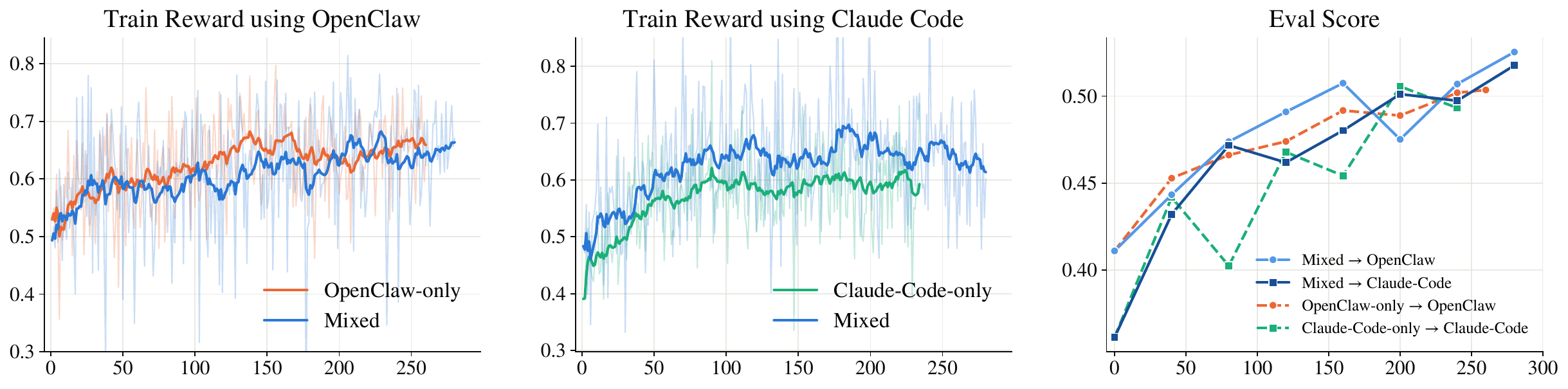}
\caption{Comparison between single-harness and mix-harness RL training.}
    \label{fig:mix-harness-training}
\end{figure}

As shown in Figure~\ref{fig:mix-harness-training}, the mixed model exhibits training rewards comparable to those of the corresponding individual-harness models. 
Under OpenClaw, it gradually narrows the initial gap toward the end of training, while under Claude Code, its training reward remains on par with or higher than that of the Claude-Code-only model.
A similar pattern is observed in downstream evaluation, where the mixed model matches or slightly outperforms the corresponding individual-harness models under both harnesses. 
Overall, combining rollouts from heterogeneous harnesses introduces no evident training instability or systematic performance degradation. 
These results demonstrate that learning signals from heterogeneous harnesses can be jointly leveraged to optimize a shared policy model within our unified black-box RL framework, without evident performance degradation under either execution system.

\subsection{Training on More Challenging Tasks}\label{sec:jobbench-and-officeqa}

To further evaluate the generality of our unified framework, we extend black-box RL to more challenging and structurally diverse task settings from JobBench~\citep{li2026jobbenchaligningagentwork} and OfficeQA~\citep{opsahlong2026officeqaproenterprisebenchmark}. JobBench targets professional workflows over heterogeneous workspaces containing diverse file formats, including images, databases, and native Office files, requiring agents to interpret and integrate information across multiple sources to complete complex tasks. OfficeQA instead emphasizes answer-centric reasoning over a large document corpus, where agents must retrieve relevant evidence, perform multi-step analysis or computation, and produce a verifiable final answer rather than a workspace artifact. Together, the two benchmarks cover substantially different forms of complex agent interaction, from artifact-oriented workspace execution to document-grounded analytical reasoning.

\begin{figure}[ht]
\centering
    \centering
    \includegraphics[width=0.80\linewidth]{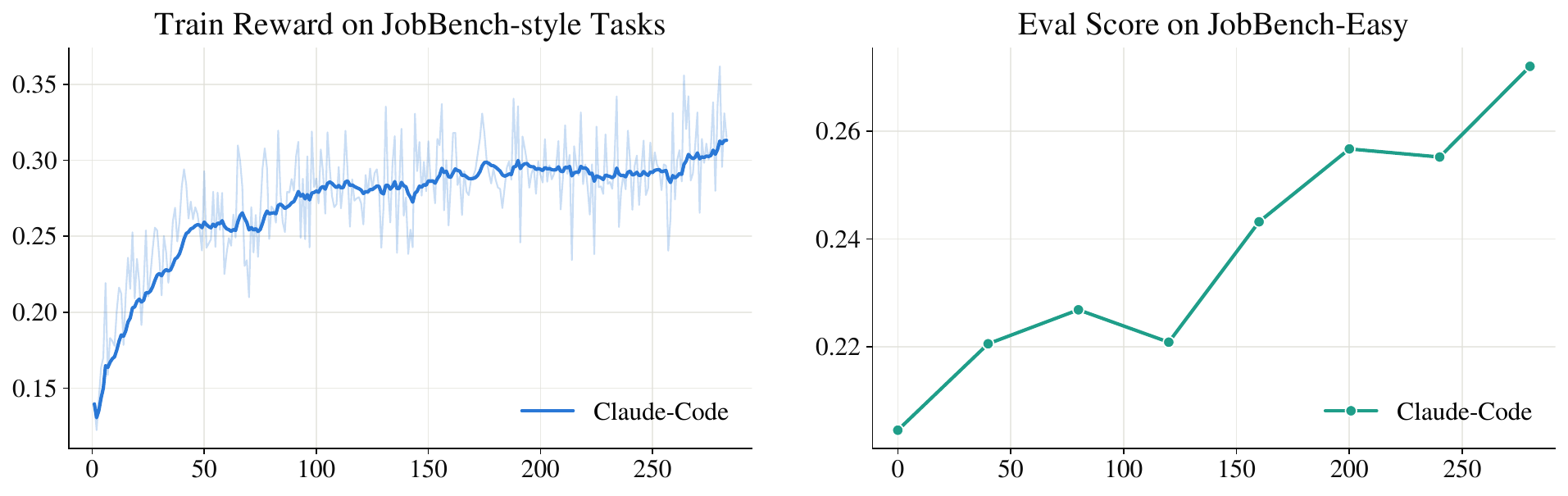}
\caption{Training dynamics with Claude Code on JobBench-style tasks.}
    \label{fig:jobbench-train}
\end{figure}

\begin{figure}[ht]
\centering
    \centering
    \includegraphics[width=0.80\linewidth]{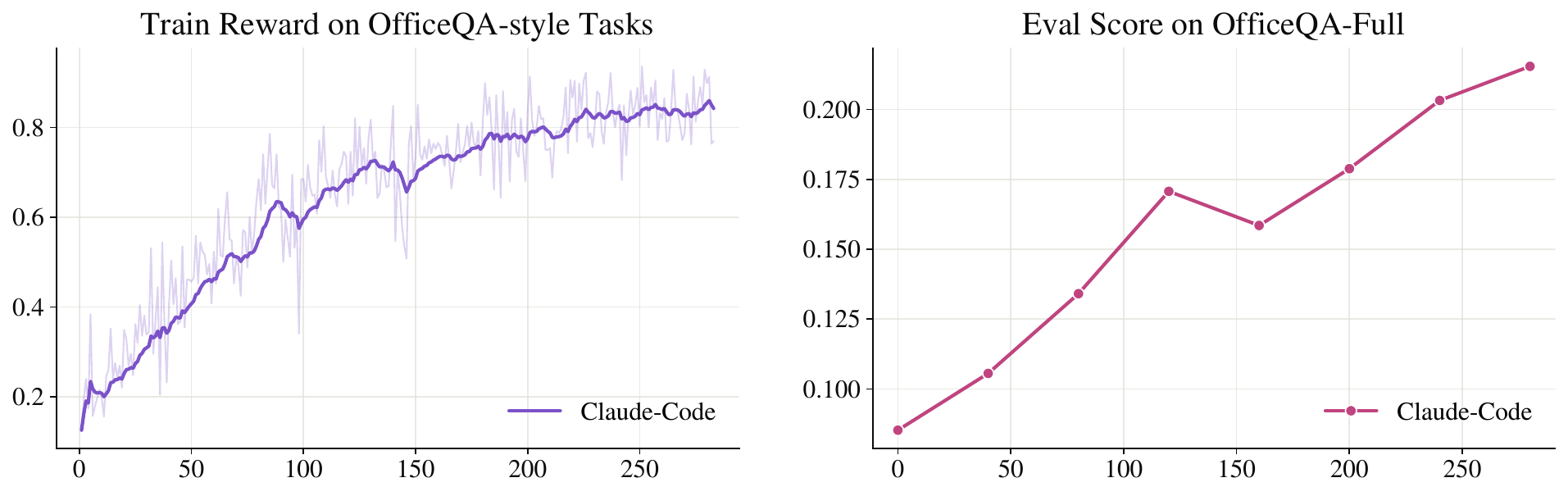}
\caption{Training dynamics with Claude Code on OfficeQA-style tasks.}
    \label{fig:officeqa-train}
\end{figure}

Following their respective task formats, we synthesize \textbf{JobBench-style} and \textbf{OfficeQA-style} training tasks and perform black-box RL on Qwen3-30A3B with Claude Code as the rollout harness. Under our unified framework, a new task can be directly incorporated into training by simply formulating it as a triplet of an instruction, an initialized workspace, and a verifier, without requiring task-specific modifications to the RL pipeline. As shown in Figures~\ref{fig:jobbench-train} and~\ref{fig:officeqa-train}, evaluation performance improves from $20.46$ to $27.20$ on JobBench-Easy and from $8.53$ to $21.54$ on OfficeQA-Full. 
Training rewards also maintain clear and stable upward trends throughout optimization. These results demonstrate that our framework is not tied to a particular task format or evaluation paradigm: the same unified black-box RL pipeline can be readily extended to substantially different and more demanding task distributions, while continuing to deliver effective and stable policy optimization.

\subsection{Effect of Cold-Start Initialization}\label{sec:cold-start}

We investigate the effect of cold-start initialization on black-box RL for general agent tasks using Qwen3-30A3B, with OpenClaw as the rollout harness and GRPO as the optimization algorithm. Specifically, we compare two initialization strategies: lightweight supervised training on a subset of trajectories from ClawGym-SynData, and direct initialization from the base model.

\begin{figure}[ht]
\centering
    \centering
    \includegraphics[width=0.95\linewidth]{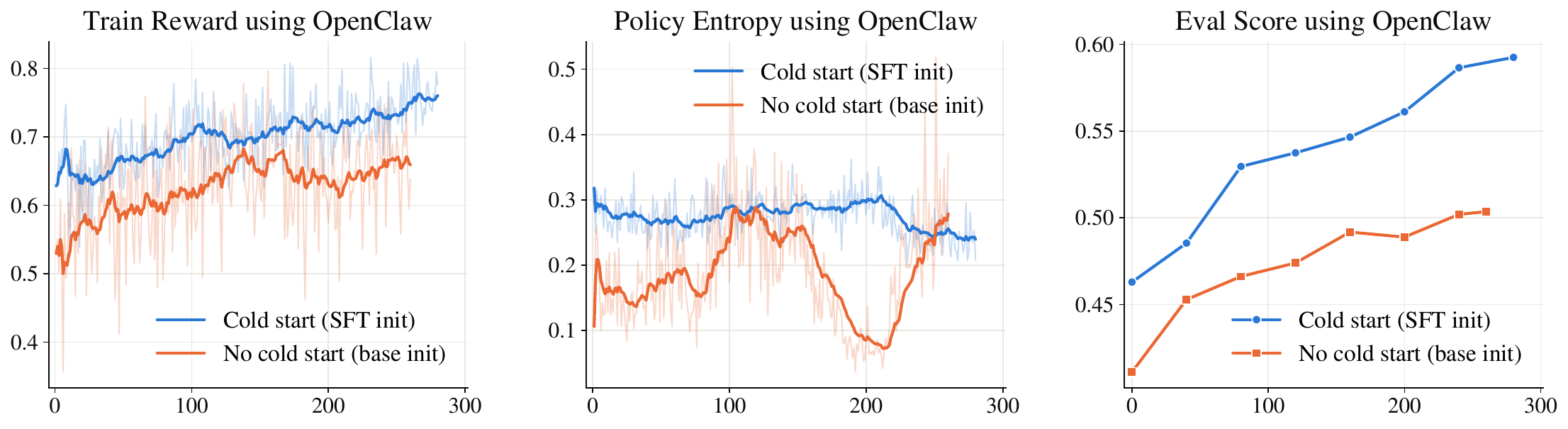}
\caption{Effect of cold-start initialization on OpenClaw black-box RL.}
    \label{fig:cold_start}
\end{figure}

As shown in Figure~\ref{fig:cold_start}, both settings benefit from RL, but the cold-started model consistently exhibits more favorable training dynamics. 
It starts from a substantially higher training reward and follows a smoother upward trajectory throughout optimization. 
Its policy entropy also remains relatively stable, whereas direct training from the base model exhibits larger fluctuations and a pronounced late-stage decline in entropy, indicating more volatile entropy dynamics and potentially less stable exploration.
Moreover, the cold-started model achieves consistently stronger downstream performance. These results indicate that lightweight cold-start training provides a better behavioral prior and improves optimization stability in this setting, although direct RL training from the base model remains effective.

\subsection{Comparison with White-Box AgentLoop RL}


Unlike black-box RL, which interacts with an opaque harness and observes only model calls at the serving boundary, white-box RL directly exposes the complete interaction process through an explicitly constructed agent loop~\citep{sun2026swe}, including the system prompt, tool interface, observation representation, context management, and workflow. These components can be independently designed and assembled to form different agent loops. 

During rollout, model generation, tool execution, observation return, and context update are directly recorded as a complete multi-turn trajectory. This makes it possible to optimize directly on the exact interaction traces produced by the agent loop.
Specifically, we instantiate a white-box agent loop with five categories of basic tools: bash, search, fetch, code interpreter, and file operations. We perform RL training through this agent loop in sandbox-based environments, and evaluate the resulting policy model under both the same white-box agent loop and an external black-box harness to assess in-loop performance and white-to-black generalization transfer. All experiments follow the same datasets and evaluation settings as the black-box counterpart, ensuring a fair comparison across training paradigms.

\begin{table*}[ht]
\caption{
Cross-harness Pass@1 performance on ClawGym-Bench under the white-box AgentLoop and OpenClaw harnesses. ClawII-OC-30A3B and WhiteBox-30A3B are trained using OpenClaw-based black-box RL and white-box AgentLoop RL, respectively. The best and second-best results within each evaluation harness are highlighted in \textbf{bold} and \underline{underlined}.
}
\centering
\footnotesize
\setlength{\tabcolsep}{3.2pt}
\renewcommand{\arraystretch}{1.12}

\begin{tabular*}{0.95\textwidth}{
@{\extracolsep{\fill}}
lccccccc
@{}
}
\toprule
\textbf{Model}
& \makecell{\textbf{Product.}\\ \textbf{\& Collab.}}
& \makecell{\textbf{Systems}\\ \textbf{\& Auto.}}
& \makecell{\textbf{Analysis}\\ \textbf{\& Reason.}}
& \makecell{\textbf{Content}\\ \textbf{\& Domain}}
& \makecell{\textbf{Planning}\\ \textbf{\& Knowl.}}
& \makecell{\textbf{Software}\\ \textbf{Dev.}}
& \textbf{Avg.} \\
\midrule

\rowcolor{gray!12}
\multicolumn{8}{c}{\textbf{White-Box AgentLoop as Harness}} \\

Qwen3-32B
& 28.79 & 24.38 & 22.34 & 27.88 & 27.25 & 26.69 & 26.43 \\

Qwen3-30A3B
& 44.22 & 35.93 & 36.85 & 39.76 & 49.55 & 48.09 & 41.69 \\

ClawII-OC-30A3B
& \underline{50.22}
& \underline{46.09}
& \underline{51.64}
& \underline{53.55}
& \underline{59.11}
& \underline{51.72}
& \underline{51.37} \\

WhiteBox-30A3B
& \textbf{62.49}
& \textbf{53.36}
& \textbf{58.75}
& \textbf{61.35}
& \textbf{68.50}
& \textbf{57.33}
& \textbf{59.90} \\

\midrule

\rowcolor{gray!12}
\multicolumn{8}{c}{\textbf{OpenClaw as Harness}} \\

Qwen3-32B
& 40.68 & 36.21 & 37.84 & 47.16 & 49.29 & 33.11 & 40.32 \\

Qwen3-30A3B
& 42.47 & 42.09 & 45.04 & 51.95 & 45.98 & 46.24 & 45.11 \\

ClawII-OC-30A3B
& \textbf{62.75}
& \textbf{59.81}
& \textbf{64.09}
& \textbf{62.18}
& \textbf{63.12}
& \textbf{66.71}
& \textbf{62.62} \\

WhiteBox-30A3B
& \underline{49.13}
& \underline{44.60}
& \underline{51.10}
& \underline{50.36}
& \underline{57.91}
& \underline{53.01}
& \underline{50.33} \\

\bottomrule
\end{tabular*}
\label{tab:whitebox_cross_harness}
\end{table*}

\begin{figure}[ht]
\centering
\includegraphics[width=0.95\linewidth]{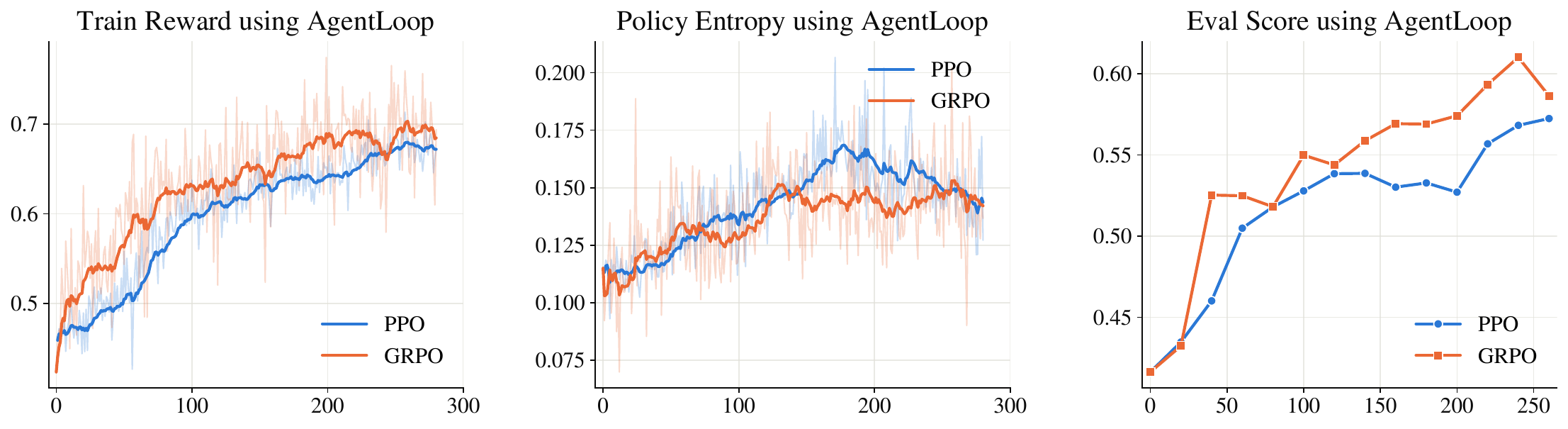}
\caption{
Training dynamics of white-box AgentLoop RL with GRPO and PPO. The curves report training reward, policy entropy, and evaluation score over the course of training.
}
\label{fig:curves-of-whitebox}
\end{figure}


\paragraph{Effectiveness under the white-box agentloop.}
White-box agentloop RL yields substantial gains when evaluated under the same interaction framework used for training. As shown in Table~\ref{tab:whitebox_cross_harness}, WhiteBox-30A3B achieves an average score of 59.90, improving over its Qwen3-30A3B initialization by 18.21 points and outperforming the black-box-trained ClawII-OC-30A3B by 8.53 points. The improvement is consistent across all six task categories. Figure~\ref{fig:curves-of-whitebox} further shows that both GRPO and PPO exhibit stable optimization dynamics under the white-box agentloop, with training reward and evaluation performance improving over the course of training. These results demonstrate that the explicitly constructed agent loop provides an effective and stable environment for policy optimization across different RL algorithms.

\paragraph{White-to-black transfer.}
The policy model trained with the white-box agentloop also transfers to the external OpenClaw harness without additional training. WhiteBox-30A3B reaches an average score of 50.33 under OpenClaw, exceeding the original Qwen3-30A3B model by 5.22 points and improving performance in five of the six categories. This transfer suggests that different harnesses share a common set of underlying agentic capabilities that can be acquired through white-box training. However, WhiteBox-30A3B still falls behind ClawII-OC-30A3B, which is trained directly under OpenClaw and achieves 62.62, revealing that training under a simple white-box agentloop may still be insufficient to fully capture the harness-specific interaction patterns required by an unseen black-box system.

%% file: sections/5_conclusion.tex
\section{Conclusion}

In this paper, we establish black-box reinforcement learning as a practical and scalable approach for training general autonomous agents directly through complex deployment harnesses. The native harness is treated as an unmodified and opaque rollout engine, thereby decoupling policy optimization from harness execution. We develop an end-to-end infrastructure that isolates rollouts in temporary sandboxes, captures faithful model behavior at the serving boundary, reconstructs forked multi-turn trajectories with prefix trees, and preserves training--inference consistency. Building on this infrastructure, we adapt both PPO and GRPO to tasks that may yield multiple trajectories and further extend the formulation to mix-harness training, enabling a single model to jointly learn from heterogeneous harnesses within a unified optimization pipeline.
Experiments on two structurally distinct harnesses, OpenClaw and Claude Code, validate the effectiveness and stability of this formulation. With Qwen3-30A3B, black-box RL improves Pass@1 by $9.98$ and $14.81$ points on ClawGym-Bench and by $11.71$ and $17.28$ points on PinchBench, while remaining stable over 200--400 optimization steps. The same framework also extends to more challenging task settings from JobBench and OfficeQA, where black-box RL continues to improve downstream performance. These results demonstrate that complex deployment harnesses can serve as effective training interfaces, providing a practical foundation for optimizing generalist agents directly within the execution systems in which they operate.

In future work, we will explore incorporating auxiliary interactions, such as compaction and subagent trajectories, into policy optimization rather than excluding them from training. We also plan to extend our framework to broader general agent tasks and more diverse execution settings to further assess its scalability and generality.

%% file: ref.bib
@misc{dressage_github,
  author       = {Liangmeng Huang and Qingchuan Li and Hongwei Xue and Shilin Yan and {Dressage Contributors}},
  title        = {{Dressage}: Scalable {RL} for Any Agent and Any Sandbox},
  year         = {2026},
  howpublished = {\url{https://github.com/Accio-Lab/Dressage}}
}

@article{xu2026polar,
  title={Polar: Agentic RL on Any Harness at Scale},
  author={Xu, Binfeng and Zhang, Hao and Zhang, Shaokun and Han, Songyang and Liu, Mingjie and Hu, Jian and Diao, Shizhe and Jin, Zhenghui and Zou, Yunheng and Demoret, Michael and others},
  journal={arXiv preprint arXiv:2605.24220},
  year={2026}
}

@article{bai2026clawgym,
  title={ClawGym: A scalable framework for building effective claw agents},
  author={Bai, Fei and Song, Huatong and Sun, Shuang and Cheng, Daixuan and Yang, Yike and Hao, Chuan and Li, Renyuan and Chang, Feng and Wei, Yuan and Tao, Ran and others},
  journal={arXiv preprint arXiv:2604.26904},
  year={2026}
}

@article{li2026kat,
  title={Kat-coder-v2 technical report},
  author={Li, Fengxiang and Zhang, Han and Huang, Haoyang and Wang, Jinghui and Hao, Jinhua and Yuan, Kun and Li, Mengtong and Zhang, Minglei and Xu, Pengcheng and Zhuang, Wenhao and others},
  journal={arXiv preprint arXiv:2603.27703},
  year={2026}
}

@misc{openai2026gpt54,
  title        = {Introducing {GPT-5.4}},
  author       = {{OpenAI}},
  year         = {2026},
  month        = mar,
  day          = {5},
  howpublished = {\url{https://openai.com/index/introducing-gpt-5-4/}},
  note         = {Official release announcement. Accessed: 2026-04-29}
}

@article{yu2026openforgerl,
  title={OpenForgeRL: Train Harness-native Agents in Any Environment},
  author={Yu, Xiao and Peng, Baolin and Xu, Ruize and Zou, Hao and Wu, Qianhui and Cheng, Hao and Yao, Wenlin and Singh, Nikhil and Yu, Zhou and Gao, Jianfeng},
  journal={arXiv preprint arXiv:2607.21557},
  year={2026}
}

@article{yang2025qwen3technicalreport,
  title={Qwen3 technical report},
  author={Yang, An and Li, Anfeng and Yang, Baosong and Zhang, Beichen and Hui, Binyuan and Zheng, Bo and Yu, Bowen and Gao, Chang and Huang, Chengen and Lv, Chenxu and others},
  journal={arXiv preprint arXiv:2505.09388},
  year={2025}
}

@misc{sweb-verified,
    title = {Introducing SWE-bench Verified},
    url = {https://openai.com/index/introducing-swe-bench-verified/},
    author = {Neil Chowdhury and James Aung and Chan Jun Shern and Oliver Jaffe and Dane Sherburn and Giulio Starace and Evan Mays and Rachel Dias and Marwan Aljubeh and Mia Glaese and Carlos E. Jimenez and John Yang and Leyton Ho and Tejal Patwardhan and Kevin Liu and Aleksander Madry},
    year={2024},
    month = {August},
}

@inproceedings{merrill2026terminalbench,
  title={Terminal-Bench: Benchmarking Agents on Hard, Realistic Tasks in Command Line Interfaces},
  author={Mike A Merrill and Alexander Glenn Shaw and Nicholas Carlini and Boxuan Li and Harsh Raj and Ivan Bercovich and Lin Shi and Jeong Yeon Shin and Thomas Walshe and E. Kelly Buchanan and Junhong Shen and Guanghao Ye and Haowei Lin and Jason Poulos and Maoyu Wang and Jenia Jitsev and Marianna Nezhurina and Di Lu and Orfeas Menis Mastromichalakis and Zhiwei Xu and Zizhao Chen and Yue Liu and Robert Zhang and Leon Liangyu Chen and Anurag Kashyap and Jan-Lucas Uslu and Jeffrey Li and Jianbo Wu and Minghao Yan and Song Bian and Vedang Sharma and Ke Sun and Steven Dillmann and Akshay Anand and Andrew Lanpouthakoun and Bardia Koopah and Changran Hu and Etash Kumar Guha and Gabriel H. S. Dreiman and Jiacheng Zhu and Karl Krauth and Li Zhong and Niklas Muennighoff and Robert Kwesi Amanfu and Shangyin Tan and Shreyas Pimpalgaonkar and Tushar Aggarwal and Xiangning Lin and Xin Lan and Xuandong Zhao and Yiqing Liang and Yuanli Wang and Zilong Wang and Changzhi Zhou and David Heineman and Hange Liu and Harsh Trivedi and John Yang and Junhong Lin and Manish Shetty and Michael Yang and Nabil Omi and Negin Raoof and Shanda Li and Terry Yue Zhuo and Wuwei Lin and Yiwei Dai and Yuxin Wang and Wenhao Chai and Shang Zhou and Dariush Wahdany and Ziyu She and Jiaming Hu and Zhikang Dong and Yuxuan Zhu and Sasha Cui and Ahson Saiyed and Arinbj{\"o}rn Kolbeinsson and Christopher Michael Rytting and Ryan Marten and Yixin Wang and Alex Dimakis and Andy Konwinski and Ludwig Schmidt},
  booktitle={The Fourteenth International Conference on Learning Representations},
  year={2026},
  url={https://openreview.net/forum?id=a7Qa4CcHak}
}

@article{song2025r1,
  title={R1-searcher: Incentivizing the search capability in llms via reinforcement learning},
  author={Song, Huatong and Jiang, Jinhao and Min, Yingqian and Chen, Jie and Chen, Zhipeng and Zhao, Wayne Xin and Fang, Lei and Wen, Ji-Rong},
  journal={arXiv preprint arXiv:2503.05592},
  year={2025}
}

@misc{officeqa,
  title = {OfficeQA: A Grounded Reasoning Benchmark},
  author = {Databricks},
  year = {2025},
  license = {CC-BY-SA-4.0}
}

@article{sun2026swe,
  title={SWE-World: Building Software Engineering Agents in Docker-Free Environments},
  author={Sun, Shuang and Song, Huatong and Huang, Lisheng and Jiang, Jinhao and Le, Ran and Lv, Zhihao and Chen, Zongchao and Hu, Yiwen and Luo, Wenyang and Zhao, Wayne Xin and others},
  journal={arXiv preprint arXiv:2602.03419},
  year={2026}
}

@article{song2026swe,
  title={SWE-Master: Unleashing the Potential of Software Engineering Agents via Post-Training},
  author={Song, Huatong and Huang, Lisheng and Sun, Shuang and Jiang, Jinhao and Le, Ran and Cheng, Daixuan and Chen, Guoxin and Hu, Yiwen and Chen, Zongchao and Jia, Yiming and others},
  journal={arXiv preprint arXiv:2602.03411},
  year={2026}
}

@misc{anthropic2026claudecode,
  title        = {{Claude Code}: AI-Powered Coding Assistant for Developers},
  author       = {{Anthropic}},
  year         = {2026},
  howpublished = {\url{https://claude.com/product/claude-code}},
  note         = {Product page. Accessed: 2026-04-29}
}

@misc{openai2026codex,
  title        = {{Codex}: AI Coding Partner from OpenAI},
  author       = {{OpenAI}},
  year         = {2026},
  howpublished = {\url{https://openai.com/codex/}},
  note         = {Product page. Accessed: 2026-04-29}
}

@misc{qwenclawbench1.1,
    title = {{QwenClawBench}: Real-user-distribution benchmark for OpenClaw agents},
    url = {github.com/SKYLENAGE-AI/QwenClawBench},
    author = {{Qwen Team} and {Data Team}, {Alibaba Group}},
    month = {April},
    year = {2026}
}

@inproceedings{yao2022react,
  title={React: Synergizing reasoning and acting in language models},
  author={Yao, Shunyu and Zhao, Jeffrey and Yu, Dian and Shafran, Izhak and Narasimhan, Karthik R and Cao, Yuan},
  booktitle={NeurIPS 2022 Foundation Models for Decision Making Workshop},
  year={2022}
}

@article{zhao2023survey,
  title={A survey of large language models},
  author={Zhao, Wayne Xin and Zhou, Kun and Li, Junyi and Tang, Tianyi and Wang, Xiaolei and Hou, Yupeng and Min, Yingqian and Zhang, Beichen and Zhang, Junjie and Dong, Zican and others},
  journal={arXiv preprint arXiv:2303.18223},
  volume={1},
  number={2},
  year={2023}
}

@article{chen2025empirical,
  title={An empirical study on eliciting and improving r1-like reasoning models},
  author={Chen, Zhipeng and Min, Yingqian and Zhang, Beichen and Chen, Jie and Jiang, Jinhao and Cheng, Daixuan and Zhao, Wayne Xin and Liu, Zheng and Miao, Xu and Lu, Yang and others},
  journal={arXiv preprint arXiv:2503.04548},
  year={2025}
}

@misc{patwardhan2025gdpvalevaluatingaimodel,
      title={GDPval: Evaluating AI Model Performance on Real-World Economically Valuable Tasks}, 
      author={Tejal Patwardhan and Rachel Dias and Elizabeth Proehl and Grace Kim and Michele Wang and Olivia Watkins and Simón Posada Fishman and Marwan Aljubeh and Phoebe Thacker and Laurance Fauconnet and Natalie S. Kim and Patrick Chao and Samuel Miserendino and Gildas Chabot and David Li and Michael Sharman and Alexandra Barr and Amelia Glaese and Jerry Tworek},
      year={2025},
      eprint={2510.04374},
      archivePrefix={arXiv},
      primaryClass={cs.LG},
      url={https://arxiv.org/abs/2510.04374}, 
}

@misc{opsahlong2026officeqaproenterprisebenchmark,
      title={OfficeQA Pro: An Enterprise Benchmark for End-to-End Grounded Reasoning}, 
      author={Krista Opsahl-Ong and Arnav Singhvi and Jasmine Collins and Ivan Zhou and Cindy Wang and Ashutosh Baheti and Owen Oertell and Jacob Portes and Sam Havens and Erich Elsen and Michael Bendersky and Matei Zaharia and Xing Chen},
      year={2026},
      eprint={2603.08655},
      archivePrefix={arXiv},
      primaryClass={cs.AI},
      url={https://arxiv.org/abs/2603.08655}, 
}

@misc{li2026jobbenchaligningagentwork,
      title={JobBench: Aligning Agent Work With Human Will}, 
      author={Yuetai Li and Yichen Feng and Zhangchen Xu and Zixian Ma and Kaiyuan Zheng and Fengqing Jiang and Xinghua Sun and Rulin Shao and Zichen Chen and Yue Huang and Xinyang Han and Brian Lee and Kayla Xu and Shenglai Zeng and Hang Hua and Xiangliang Zhang and Basel Alomair and Ranjay Krishna and Luke Zettlemoyer and Pang Wei Koh and Bhaskar Ramasubramanian and Luyao Niu and Xiang Yue and Radha Poovendran},
      year={2026},
      eprint={2605.26329},
      archivePrefix={arXiv},
      primaryClass={cs.AI},
      url={https://arxiv.org/abs/2605.26329}, 
}

@article{ye2026claw,
  title={Claw-Eval: Toward Trustworthy Evaluation of Autonomous Agents},
  author={Ye, Bowen and Li, Rang and Yang, Qibin and Liu, Yuanxin and Yao, Linli and Lv, Hanglong and Xie, Zhihui and An, Chenxin and Li, Lei and Kong, Lingpeng and others},
  journal={arXiv preprint arXiv:2604.06132},
  year={2026}
}

@misc{tang2026agent,
  title={Agent Systems with Harness Engineering},
  author={Xinyu Tang and Han Peng and Guoxin Chen and Yuze Shi and
          Zitao Su and Peiyu Liu and Wayne Xin Zhao and Yawen Li and
          Zhe Xue},
  url={https://openreview.net/pdf?id=nM5tDHrQsx},
  year={2026}
}

@misc{openclaw2026,
  title        = {{OpenClaw}: Personal AI Assistant},
  author       = {{OpenClaw}},
  year         = {2026},
  howpublished = {\url{https://github.com/openclaw/openclaw}},
  note         = {GitHub repository. Accessed: 2026-04-29}
}

@misc{pinchbench2026,
  title        = {{PinchBench}: Real-World Benchmarks for AI Coding Agents},
  author       = {{PinchBench}},
  year         = {2026},
  howpublished = {\url{https://github.com/pinchbench/skill}},
  note         = {GitHub repository. Accessed: 2026-04-29}
}

@misc{qwenpaw2026,
  title        = {{QwenPaw}: Your Personal AI Assistant},
  author       = {{AgentScope-AI}},
  year         = {2026},
  howpublished = {\url{https://github.com/agentscope-ai/QwenPaw}},
  note         = {GitHub repository. Accessed: 2026-04-29}
}

@misc{nanobot2026,
  title        = {{NanoBot}: The Ultra-Lightweight Personal AI Agent},
  author       = {{HKUDS}},
  year         = {2026},
  howpublished = {\url{https://github.com/HKUDS/nanobot}},
  note         = {GitHub repository. Accessed: 2026-04-29}
}

@misc{yao2025offpolicy,
  title = {Your Efficient RL Framework Secretly Brings You Off-Policy RL Training},
  url = {https://fengyao.notion.site/off-policy-rl},
  author = {Yao, Feng and Liu, Liyuan and Zhang, Dinghuai and Dong, Chengyu and Shang, Jingbo and Gao, Jianfeng},
  journal = {Feng Yao's Notion},
  year = {2025},
  month = aug,
}

@misc{liu-li-2025-rl-collapse,
  title = {When Speed Kills Stability: Demystifying {RL} Collapse from the Training-Inference Mismatch},
  author = {Liu, Jiacai and Li, Yingru and Fu, Yuqian and Wang, Jiawei and Liu, Qian and Shen, Yu},
  year = {2025},
  month = sep,
  url = {https://richardli.xyz/rl-collapse}
}

@article{wang2026harness,
  title={Harness Handbook: Making Evolving Agent Harnesses Readable, Navigable, and Editable},
  author={Wang, Ruhan and Shi, Yucheng and Li, Zongxia and Li, Zhongzhi and Yu, Yue and Yang, Junyao and Panaganti, Kishan and Mi, Haitao and Zhou, Dongruo and others},
  journal={arXiv preprint arXiv:2607.13285},
  year={2026}
}

@article{deepseekmath,
  title={Deepseekmath: Pushing the limits of mathematical reasoning in open language models},
  author={Shao, Zhihong and Wang, Peiyi and Zhu, Qihao and Xu, Runxin and Song, Junxiao and Bi, Xiao and Zhang, Haowei and Zhang, Mingchuan and Li, YK and Wu, Yang and others},
  journal={arXiv preprint arXiv:2402.03300},
  year={2024}
}

@article{schulman2017proximal,
  title={Proximal policy optimization algorithms},
  author={Schulman, John and Wolski, Filip and Dhariwal, Prafulla and Radford, Alec and Klimov, Oleg},
  journal={arXiv preprint arXiv:1707.06347},
  year={2017}
}

@misc{gallouedec2026tito,
  title   = {Agentic RL: Token-In, Token-Out Done Right},
  author  = {Gallou{\'e}dec, Quentin and Rasul, Kashif},
  year    = {2026},
  month   = may,
  day     = {29},
  url     = {https://huggingface.co/blog/huggingface/tito},
  urldate = {2026-07-28},
  note    = {Hugging Face Blog}
}

@article{li2025tool,
  title={The tool decathlon: Benchmarking language agents for diverse, realistic, and long-horizon task execution},
  author={Li, Junlong and Zhao, Wenshuo and Zhao, Jian and Zeng, Weihao and Wu, Haoze and Wang, Xiaochen and Ge, Rui and Cao, Yuxuan and Huang, Yuzhen and Liu, Wei and others},
  journal={arXiv preprint arXiv:2510.25726},
  year={2025}
}

@article{bandi2026mcp,
  title={Mcp-atlas: A large-scale benchmark for tool-use competency with real mcp servers},
  author={Bandi, Chaithanya and Dumitru, Razvan-Gabriel and Hertzberg, Ben and Agarwal, Divyansh and Boo, Geobio and Polakam, Tejas and Hassaan, Sami and Da, Jeff and Kim, HiJae and Gupta, Vipul and others},
  journal={arXiv preprint arXiv:2602.00933},
  year={2026}
}

@misc{rucnlpir2026awesomelonghorizonagents,
  title        = {Awesome Long-Horizon Agents},
  author       = {{RUC-NLPIR}},
  year         = {2026},
  howpublished = {\url{https://github.com/RUC-NLPIR/Awesome-Long-Horizon-Agents}},
  note         = {GitHub repository},
  urldate      = {2026-07-28}
}

@inproceedings{bai2026towards,
  title={Towards effective code-integrated reasoning},
  author={Bai, Fei and Min, Yingqian and Zhang, Beichen and Chen, Zhipeng and Zhao, Xin and Fang, Lei and Liu, Zheng and Wang, Zhongyuan and Xu, Hongteng},
  booktitle={Proceedings of the AAAI Conference on Artificial Intelligence},
  volume={40},
  pages={30022--30030},
  year={2026}
}
